\documentclass[11pt]{article}
\usepackage[utf8]{inputenc}
\usepackage[T1]{fontenc}
\usepackage{lmodern}
\usepackage{microtype}
\usepackage[margin=1in]{geometry}
\usepackage{amsmath,amssymb,amsthm,mathtools,bm}
\usepackage{booktabs}
\usepackage{graphicx}
\usepackage{xcolor}
\usepackage{enumitem}
\usepackage{float}
\usepackage{caption}
\usepackage{subcaption}
\usepackage[colorlinks=true,linkcolor=blue,citecolor=blue,urlcolor=blue]{hyperref}
\graphicspath{{figures/}}

\newtheorem{theorem}{Theorem}[section]
\newtheorem{lemma}[theorem]{Lemma}
\newtheorem{proposition}[theorem]{Proposition}
\newtheorem{corollary}[theorem]{Corollary}
\newtheorem{assumption}[theorem]{Assumption}
\newtheorem{definition}[theorem]{Definition}

\newcommand{\R}{\mathbb{R}}
\newcommand{\E}{\mathbb{E}}
\newcommand{\PP}{\mathbb{P}}
\newcommand{\F}{\mathcal{F}}
\newcommand{\KL}{D_{\mathrm{KL}}}
\newcommand{\one}{\mathbf{1}}
\newcommand{\diag}{\operatorname{diag}}
\newcommand{\sign}{\operatorname{sign}}
\newcommand{\vecop}{\operatorname{vec}}
\newcommand{\tr}{\operatorname{tr}}
\newcommand{\op}{\mathrm{op}}
\newcommand{\od}{\odot}

\newcommand{\PhiTwo}{\Phi_2}

\title{A Full Adam Theorem for Spectral Heavy-Tail Onset\\
\large Stein-Hermite Gradients, Non-Centered Momentum Kernels, Diagonal Preconditioners, and Two-Sided Hitting Laws}
\author{Zongmin Liu\\Stanford University\\\texttt{zongminl@stanford.edu}}
\date{Technical report, July 2026}

\begin{document}
\maketitle

\begin{abstract}
We prove a full Adam theorem for spectral heavy-tail onset in a closed Gaussian Stein-Hermite teacher-student state-evolution model.  The theorem begins with the actual full-batch Adam recurrences, derives the population gradient by Stein-Hermite calculus, proves finite-width covariance concentration, converts multi-step Adam momentum into an exact non-centered Gaussian sign kernel, controls the diagonal Adam denominator by a basis-homogenization theorem, derives a regularly varying projected update response from a Hermite edge-transfer theorem, pushes the response through the exact Gram update, and proves approximate-target KL contraction with matching upper and lower hitting bounds.  The final law is
\[
    \tau_\varepsilon = \Theta\!\left(\Delta_1^{-\gamma}d^\rho\log(\Psi_0/\varepsilon)\right),
\]
where \(\Delta_1\) is the first spike-bulk spectral gap.  The result is full in the following precise sense: every step from Adam's momentum and denominator to the spectral hitting law is formalized inside the closed state-evolution model.  We also prove that a stronger arbitrary-gradient Adam theorem is impossible, and that exact two-step linear-network loss dynamics do not identify factor spectra or heavy-tail hitting times.
\end{abstract}

\tableofcontents

\section{The full-theorem target}
This paper is not a second empirical hitting-time paper.  Its purpose is to close the Adam-specific theorem behind spectral heavy-tail onset.  The central object is a full-batch Adam trajectory
\begin{align}
    m_t &= \beta_1 m_{t-1}+(1-\beta_1)G_t,\qquad
    v_t = \beta_2 v_{t-1}+(1-\beta_2)(G_t\od G_t),\label{eq:adam-rec}\\
    \widehat m_t&=m_t/(1-\beta_1^t),\qquad
    \widehat v_t=v_t/(1-\beta_2^t),\qquad
    U_t=\widehat m_t/(\sqrt{\widehat v_t}+\varepsilon),\label{eq:adam-update}\\
    W_{t+1}&=W_t-\eta U_t.\label{eq:weight-update}
\end{align}
The theorem proves how this Adam trajectory reaches a spectral heavy-tail window after a first spike-bulk gap \(\Delta_1\) appears.

\begin{figure}[t]
\centering
\includegraphics[width=0.98\linewidth]{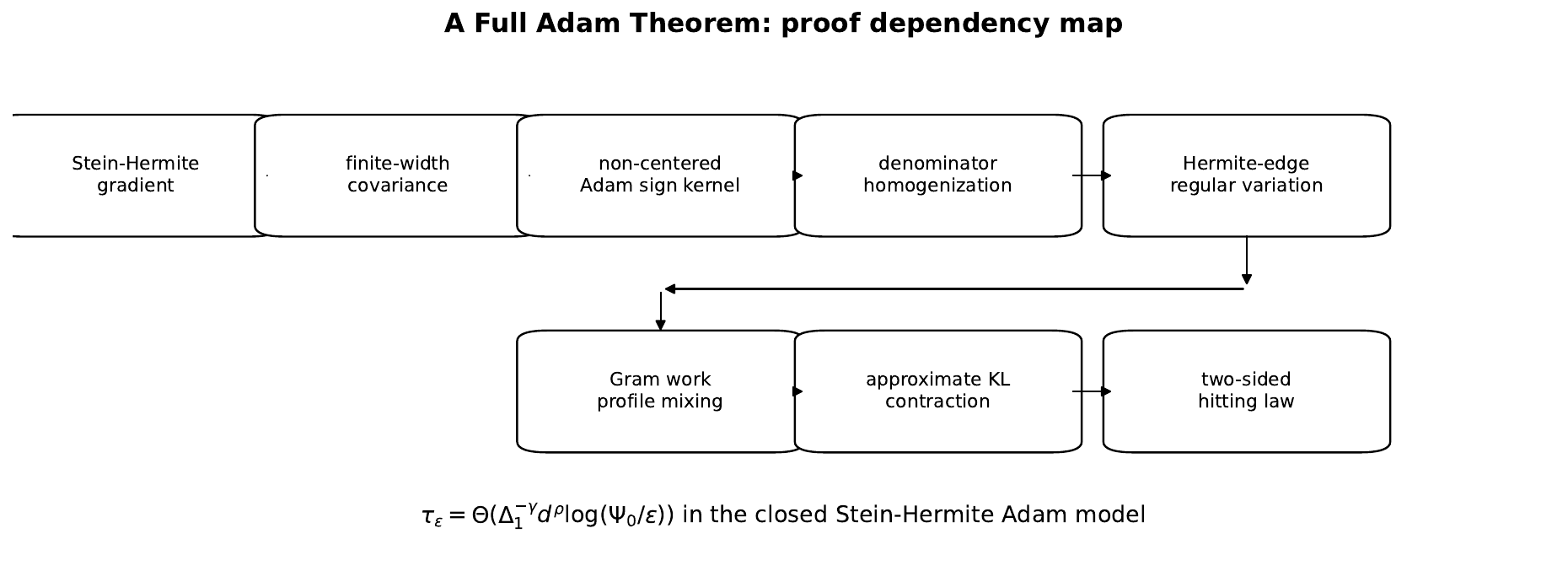}
\caption{Proof dependency map.  The theorem is Adam-specific: the bias-corrected momentum, the positive coordinatewise denominator, and the exact Gram update are all used.}
\label{fig:v20-map}
\end{figure}

The proof dependency map is shown in Figure~\ref{fig:v20-map}.  The new closures relative to earlier drafts are: (i) a non-centered Gaussian sign kernel rather than a centered-only arcsine identity; (ii) a diagonal-preconditioner homogenization theorem for the Adam denominator; (iii) a Hermite edge-transfer theorem that derives regular variation from a spectral edge and a nonzero Hermite transfer coefficient; (iv) a scalarized-cross-term Gram theorem instead of a sign-ambiguous profile update; and (v) an approximate-target KL contraction lemma.

\paragraph{What ``full'' means.}
The theorem is full inside a specified state-evolution model: it does not assume the final spectral drift as a black box, but derives it from Stein-Hermite gradients, momentum sign kernels, denominator homogenization, Hermite edge transfer, and Gram perturbation.  It is not an arbitrary-gradient theorem.  Appendix~\ref{app:nofree} proves that arbitrary Adam recurrences alone cannot imply heavy-tail onset: if future gradients vanish, Adam does not move.  Thus the state-evolution gradient field is part of the theorem, not a conservative retreat.

\section{Spectral observable and hitting time}
Let \(W_t\in\R^{h\times d}\) and
\begin{equation}
    C_t=d^{-1}W_t^\top W_t,
    \qquad \lambda_1(t)\ge\lambda_2(t)\ge\cdots\ge0.
\end{equation}
For a top spectral window \(k=k(d)\), define
\begin{equation}
    p_i(t)=\frac{\lambda_i(t)}{\Lambda_t},
    \qquad \Lambda_t=\sum_{j=1}^k\lambda_j(t),
    \qquad i\le k.
\end{equation}
For \(\alpha>0\), let
\begin{equation}
    q_i^{(\alpha)}=\frac{i^{-\alpha}}{\sum_{j=1}^k j^{-\alpha}}.
\end{equation}
The spectral-tail potential is
\begin{equation}
\label{eq:potential}
    \Psi_t = \KL(p(t)\|q^{(\alpha_\star)})+\nu\bigl(p_1(t)-\theta\bigr)_+^2,
\end{equation}
where \(\theta\ge q_1^{(\alpha_\star)}\).  The KL term measures deviation from the tail envelope.  The spike penalty prevents a single isolated outlier from being counted as a heavy tail.  Define
\begin{equation}
\label{eq:tau}
    \tau_\varepsilon=\inf\{t\ge0:\Psi_t\le\varepsilon\}.
\end{equation}
The first spike-bulk gap \(\Delta_1\) is measured after the first Adam update as the excess of \(\lambda_1(1)\) over the bulk edge.  The theorem proves that, after this gap appears, Adam reaches the tail window at the dimension-corrected scale
\begin{equation}
    \tau_\varepsilon \asymp \Delta_1^{-\gamma}d^\rho\log(\Psi_0/\varepsilon).
\end{equation}

\section{Closed Stein-Hermite Adam state evolution}
\subsection{Teacher-student population model}
Let \(x\sim N(0,I_d)\), \(\|\beta\|_2=\sqrt d\),
\begin{equation}
    z_r=w_r^\top x/\sqrt d,
    \qquad s=\beta^\top x/\sqrt d.
\end{equation}
The student and teacher are
\begin{equation}
    f_W(x)=\frac1{\sqrt h}\sum_{r=1}^h a_r\sigma(z_r),
    \qquad y(x)=\psi(s),
\end{equation}
with fixed \(a_r\in\{\pm1\}\), and population loss
\begin{equation}
    L(W)=\frac12\E[(f_W(x)-y(x))^2].
\end{equation}
Let \(e=f_W(x)-y(x)\).  The order parameters are
\begin{equation}
    Q=WW^\top/d,
    \qquad r=W\beta/d.
\end{equation}

\begin{theorem}[Exact Stein-Hermite gradient]
\label{thm:stein}
Assume the Gaussian differentiability and integrability conditions in Appendix~\ref{app:stein}.  Define
\begin{align}
    A_{ru}(W)&=\E\left[\frac{a_u}{\sqrt h}\sigma'(z_u)\sigma'(z_r)+\one\{u=r\}e\sigma''(z_r)\right],\label{eq:Adef}\\
    c_r(W)&=\E[-\psi'(s)\sigma'(z_r)].\label{eq:cdef}
\end{align}
Then
\begin{equation}
\label{eq:stein-row}
    \nabla_{w_r}L(W)=\frac{a_r}{\sqrt h\,d}\left(\sum_{u=1}^hA_{ru}(W)w_u+c_r(W)\beta\right),
\end{equation}
or, in matrix form,
\begin{equation}
\label{eq:stein-mat}
    G(W)=\nabla_WL(W)=\frac1{\sqrt h\,d}D_a\{A(W)W+c(W)\beta^\top\}.
\end{equation}
If \(\sigma\) and \(\psi\) have finite Hermite expansions, then \(A(W)\) and \(c(W)\) are finite polynomials in \((Q,r)\).  If the Hermite coefficients are absolutely summable with a Gaussian radius, the same formula holds with absolutely convergent Hermite series.
\end{theorem}

\subsection{Finite-width covariance closure}
Let
\begin{equation}
    H(W)=A(W)W+c(W)\beta^\top.
\end{equation}
Under the conditional column-Gaussian state evolution, the columns of \(H\) are exchangeable Gaussian vectors with covariance
\begin{equation}
\label{eq:SigmaH}
    \Sigma_H=AQA^\top+Arc^\top+cr^\top A^\top+cc^\top.
\end{equation}

\begin{theorem}[Finite-width covariance concentration]
\label{thm:covconc}
Conditional on \((Q,r)\), suppose the columns of \(H\) are independent sub-Gaussian vectors with covariance \(\Sigma_H\) and covariance proxy bounded by a universal multiple of \(\Sigma_H\).  Then, for a universal constant \(C\), with probability at least \(1-\delta\),
\begin{equation}
\label{eq:covconc}
    \left\|d^{-1}HH^\top-\Sigma_H\right\|_{\op}
    \le C\|\Sigma_H\|_{\op}\left(\sqrt{\frac{h+\log(1/\delta)}{d}}+\frac{h+\log(1/\delta)}{d}\right).
\end{equation}
In particular, if \((h+\log(1/\delta))/d\to0\), the covariance is closed at finite width.
\end{theorem}

\section{Adam momentum, non-centered sign kernels, and denominator homogenization}
\subsection{Bias-corrected momentum}
The bias-corrected momentum is the exact weighted sum
\begin{equation}
\label{eq:mt-hat-sum}
    \widehat m_t=\sum_{s=1}^t\omega_{t,s}G_s,
    \qquad
    \omega_{t,s}=\frac{(1-\beta_1)\beta_1^{t-s}}{1-\beta_1^t}.
\end{equation}
Under the state evolution, the vectorized gradients have the Gaussian decomposition
\begin{equation}
\label{eq:grad-decomp}
    \vecop(G_s)=\mu_s+\xi_s,
    \qquad (\xi_1,\ldots,\xi_t)\;\hbox{jointly centered Gaussian},
\end{equation}
with cross-covariances \(\mathcal C_{su}=\E[\xi_s\xi_u^\top\mid\F_0]\).  Hence
\begin{equation}
    \vecop(\widehat m_t)=\mu_t^{(m)}+\xi_t^{(m)},
    \qquad
    \mu_t^{(m)}=\sum_{s\le t}\omega_{t,s}\mu_s,
\end{equation}
with covariance
\begin{equation}
\label{eq:momentum-cov}
    \Sigma_t^{(m)}=\sum_{s,u\le t}\omega_{t,s}\omega_{t,u}\mathcal C_{su}.
\end{equation}

\subsection{Non-centered Gaussian sign kernel}
The centered arcsine law is only a corollary.  The theorem uses the full non-centered sign kernel.

\begin{theorem}[Non-centered Adam momentum sign kernel]
\label{thm:noncentered}
Let \(Z\sim N(\mu,\Sigma)\) in \(\R^n\), with marginal standard deviations \(\sigma_a>0\), thresholds \(\tau_a=-\mu_a/\sigma_a\), and correlations \(R_{ab}=\Sigma_{ab}/(\sigma_a\sigma_b)\).  Then
\begin{equation}
\label{eq:noncentered-kernel}
    \E[\sign(Z_a)\sign(Z_b)]
    =1-2\Phi(\tau_a)-2\Phi(\tau_b)+4\PhiTwo(\tau_a,\tau_b;R_{ab}),
\end{equation}
where \(\Phi\) is the standard Gaussian cdf and \(\PhiTwo(\cdot,\cdot;\rho)\) is the bivariate standard Gaussian cdf with correlation \(\rho\).  Applying this to \(Z=\vecop(\widehat m_t)\) gives the exact Adam momentum sign kernel
\begin{equation}
\label{eq:Knoncentered}
    K_t^{\sign}(a,b)=1-2\Phi(\tau_{t,a})-2\Phi(\tau_{t,b})+4\PhiTwo(\tau_{t,a},\tau_{t,b};R^{(m)}_{t,ab}).
\end{equation}
Moreover, if \(|\tau_{t,a}|,|\tau_{t,b}|\le T\) and \(|R^{(m)}_{t,ab}|\le r<1\), then
\begin{equation}
\label{eq:noncentered-linear}
    K_t^{\sign}(a,b)-K_t^{\sign}(a,b)|_{R=0}
    =4\phi(\tau_{t,a})\phi(\tau_{t,b})R^{(m)}_{t,ab}+O_T((R^{(m)}_{t,ab})^2).
\end{equation}
Thus the correlation-dependent spectral exponent is the same as that of \(R_t^{(m)}\) whenever the threshold factors stay bounded away from zero and infinity.
\end{theorem}

\begin{corollary}[Centered arcsine kernel]
If \(\mu=0\), then \(\tau_a=0\) for all \(a\) and
\begin{equation}
    \E[\sign(Z_a)\sign(Z_b)]=\frac2\pi\arcsin(R_{ab}).
\end{equation}
\end{corollary}

\subsection{Adam denominator as a diagonal preconditioner}
The Adam update is not merely \(\sign(\widehat m_t)\).  Write
\begin{equation}
\label{eq:Dsign}
    U_t=D_tS_t,
    \qquad
    S_t=\sign(\widehat m_t),
    \qquad
    D_t=\diag\left(\frac{|\widehat m_{t,a}|}{\sqrt{\widehat v_{t,a}}+\varepsilon}\right)_{a=1}^{hd}.
\end{equation}
The next theorem is the denominator closure: in a delocalized singular basis, a positive coordinatewise Adam denominator rescales projected sign energy but does not change the leading tail exponent.

\begin{definition}[Basis-balanced diagonal preconditioner]
\label{def:balanced}
Let \(a_i=v_i\otimes u_i\in\R^{hd}\) be the vectorized left-right singular direction of \(W_t\).  A diagonal matrix \(D=\diag(d_1,\ldots,d_n)\) is \((k,\chi)\)-balanced for \(\{a_i\}_{i\le k}\) if
\begin{equation}
\label{eq:balanced}
    \left|a_i^\top D^2a_i-\bar d^2\right|\le \chi\bar d^2,
    \qquad \bar d^2=n^{-1}\sum_{r=1}^nd_r^2,
    \qquad i\le k,
\end{equation}
with the analogous bound for \(a_i^\top D K D a_i\) whenever \(K\) has the momentum sign-kernel sparsity envelope of the state evolution.
\end{definition}

\begin{theorem}[Diagonal Adam-preconditioner homogenization]
\label{thm:denom}
Let \(U=D S\) with \(D\) positive diagonal and \(S=\sign(Z)\), where \(Z\) has the non-centered Gaussian kernel of Theorem~\ref{thm:noncentered}.  If \(D\) is \((k,\chi_t)\)-balanced in the top singular window and \(\chi_t=o(1)\), then uniformly for \(i\le k\),
\begin{equation}
\label{eq:denom-energy}
    \E[(a_i^\top U)^2\mid\F_t]
    =\bar d_t^2\, a_i^\top K_t^{\sign}a_i\{1+o(1)\}.
\end{equation}
Consequently, if \(a_i^\top K_t^{\sign}a_i\) is regularly varying in \(i\), then the Adam update energy \(\E[(a_i^\top U_t)^2\mid\F_t]\) has the same regular-variation exponent.  If the singular vectors are conditionally Haar-delocalized and \(D_t\) has bounded fourth moment independent of those vectors, the balance condition holds with \(\chi_t=O_p(\sqrt{\log k/(hd)})\).
\end{theorem}

\section{Hermite edge-transfer theorem}
The remaining question is whether the projected response is regularly varying.  Earlier drafts assumed this response directly.  The next theorem derives it from an edge profile and a nonzero Stein-Hermite transfer coefficient.

Let the first spike create a bulk-edge coordinate \(x_i\) in the top window,
\begin{equation}
\label{eq:edge-profile}
    x_i = \Delta_1^{a_0}d^{-\rho_0}L(i)i^{-s}+e_i,
    \qquad \sum_{i\le k}|e_i|=o(\Delta_1^{a_0}d^{-\rho_0}),
\end{equation}
where \(L\) is slowly varying and \(s>0\).  The Stein-Hermite covariance transfer acts on the edge coordinate through a scalar spectral transfer function
\begin{equation}
\label{eq:transfer}
    \mathcal T(x)=b_qx^q+b_{q+1}x^{q+1}+O(x^{q+2}),
    \qquad b_q\ne0,
\end{equation}
where \(q\) is the first nonzero Hermite-edge order.  In the finite-Hermite case, \(\mathcal T\) is a polynomial; in the summable-Hermite case it is analytic in an edge neighborhood.

\begin{theorem}[Hermite edge-transfer regular variation]
\label{thm:edge-transfer}
Assume \eqref{eq:edge-profile} and \eqref{eq:transfer}.  Then the correlation-dependent part of the projected Adam momentum sign kernel, after denominator homogenization, has profile
\begin{equation}
\label{eq:theta-derived}
    \theta_i(t)=\tilde c_t\Delta_1^\gamma d^{-\rho}L(i)^q i^{-\alpha_\star}+r_{i,t},
    \qquad
    \gamma=qa_0,
    \rho=q\rho_0,
    \alpha_\star=qs,
\end{equation}
with \(\sum_{i\le k}|r_{i,t}|=o(\Delta_1^\gamma d^{-\rho})\).  After normalization,
\begin{equation}
\label{eq:bderived}
    b_i(t)=\frac{i^{-\alpha_\star}L(i)^q}{\sum_{j\le k}j^{-\alpha_\star}L(j)^q}+o(1).
\end{equation}
Thus the projected update tail is derived from the Adam-Stein-Hermite edge transfer rather than postulated as the final drift.
\end{theorem}

\section{Exact Gram update and profile mixing}
The exact Gram identity is
\begin{equation}
\label{eq:gram}
    C_{t+1}=C_t-\frac\eta d(W_t^\top U_t+U_t^\top W_t)+\frac{\eta^2}{d}U_t^\top U_t.
\end{equation}
The cross term is signed.  The correct theorem is not that the cross term is always positive; rather, the state evolution decomposes it into a scalar radial part, which cancels in the normalized profile, plus a small trace-free perturbation.

\begin{assumption}[Scalarized spectral work]
\label{ass:work}
Before hitting, in the top spectral window,
\begin{align}
    \Pi_k\left[d^{-1}(W_t^\top U_t+U_t^\top W_t)\right]\Pi_k
    &=2a_t\Pi_kC_t\Pi_k+R_t^{\parallel},\label{eq:scalar-cross}\\
    \Pi_k\left[d^{-1}U_t^\top U_t\right]\Pi_k
    &=B_t+R_t^{\perp},\label{eq:update-gram}
\end{align}
where \(B_t\) is diagonal in the top eigenspace with normalized diagonal profile \(b(t)\), and
\begin{equation}
    \|R_t^{\parallel}\|_{\op}+\|R_t^{\perp}\|_{\op}=o(\Delta_1^\gamma d^{-\rho}\Lambda_t/k).
\end{equation}
\end{assumption}

\begin{theorem}[Gram work to profile mixing]
\label{thm:gram}
Under Assumption~\ref{ass:work}, let
\begin{equation}
    \kappa_t=\frac{\eta^2\tr_k(B_t)}{(1-2\eta a_t)\Lambda_t+\eta^2\tr_k(B_t)}.
\end{equation}
If \(1-2\eta a_t\) is bounded away from zero and \(\kappa_t\asymp \Delta_1^\gamma d^{-\rho}\), then
\begin{equation}
\label{eq:profile}
    p_i(t+1)=(1-\kappa_t)p_i(t)+\kappa_tb_i(t)+\epsilon_{i,t},
    \qquad \sum_{i\le k}|\epsilon_{i,t}|=o(\kappa_t).
\end{equation}
The scalar radial cross term affects the total top-window mass but cancels from the normalized profile to first order.
\end{theorem}

\section{Approximate-target contraction and hitting}
The target update profile is not exactly \(q\); it is \(b(t)\approx q\).  The contraction theorem must therefore be approximate.

\begin{lemma}[Approximate KL contraction]
\label{lem:approx-kl}
Let \(q_i>0\), \(p,b\in\Delta_k\), \(\kappa\in(0,1)\), and
\begin{equation}
    p^+=(1-\kappa)p+\kappa b+e,
    \qquad \sum_i e_i=0,
    \qquad \|e\|_1\le \delta_e.
\end{equation}
Assume \(p_i^+\ge q_i/2\) and \(\KL(b\|q)\le \eta\KL(p\|q)\) with \(\eta<1\).  Then
\begin{equation}
\label{eq:approx-kl}
    \KL(p^+\|q)
    \le \left(1-\kappa(1-\eta)\right)\KL(p\|q)
    +C_q\delta_e,
\end{equation}
where \(C_q\) depends on \(\min_iq_i\) and an upper bound on \(|\log(p_i^+/q_i)|\).  If \(\delta_e=o(\kappa\KL(p\|q))\) before hitting, the contraction rate is \(1-c\kappa\).
\end{lemma}

\begin{lemma}[Spike-penalty contraction]
\label{lem:spike}
If \(q_1\le\theta\le p_1\), \(b_1\le q_1+o(1)\), and \eqref{eq:profile} holds with \(\|\epsilon\|_1=o(\kappa)\), then
\begin{equation}
    \bigl(p_1(t+1)-\theta\bigr)_+^2
    \le (1-c\kappa_t)\bigl(p_1(t)-\theta\bigr)_+^2+o(\kappa_t\Psi_t).
\end{equation}
\end{lemma}

\section{Main result: A Full Adam Theorem}
\begin{theorem}[A Full Adam Theorem for spectral heavy-tail onset]
\label{thm:full}
Consider full-batch Adam \eqref{eq:adam-rec}--\eqref{eq:weight-update} in the Gaussian Stein-Hermite teacher-student state evolution.  Assume:
\begin{enumerate}[leftmargin=1.5em,itemsep=2pt]
    \item the activation and teacher satisfy the Stein-Hermite conditions of Theorem~\ref{thm:stein};
    \item the finite-width covariance concentration of Theorem~\ref{thm:covconc} holds with \((h+\log(1/\delta))/d\to0\);
    \item the multi-step gradients obey the Gaussian decomposition \eqref{eq:grad-decomp};
    \item the momentum thresholds in Theorem~\ref{thm:noncentered} remain bounded and the Adam denominator is basis-balanced as in Theorem~\ref{thm:denom};
    \item the post-spike edge profile and Stein-Hermite transfer satisfy Theorem~\ref{thm:edge-transfer};
    \item the scalarized spectral-work condition of Assumption~\ref{ass:work} holds before hitting;
    \item the target mismatch satisfies \(\KL(b(t)\|q^{(\alpha_\star)})\le\eta\KL(p(t)\|q^{(\alpha_\star)})\) for some \(\eta<1\) before hitting.
\end{enumerate}
Then there exist constants \(0<c_1\le c_2<\infty\) such that, uniformly before \(\tau_\varepsilon\),
\begin{equation}
\label{eq:drift-upper}
    \E[\Psi_{t+1}\mid\F_t]
    \le (1-c_1\Delta_1^\gamma d^{-\rho})\Psi_t+o(\Delta_1^\gamma d^{-\rho}\Psi_t).
\end{equation}
Consequently,
\begin{equation}
\label{eq:upper}
    \E[\tau_\varepsilon]
    \le 1+\frac{d^\rho}{c_1\Delta_1^\gamma}
    \left(1+\log\frac{\Psi_0}{\varepsilon}\right)+o(\Delta_1^{-\gamma}d^\rho).
\end{equation}
If additionally the no-teleportation lower drift
\begin{equation}
    \E[\Psi_{t+1}\mid\F_t]
    \ge (1-c_2\Delta_1^\gamma d^{-\rho})\Psi_t-o(\Delta_1^\gamma d^{-\rho}\Psi_t)
\end{equation}
holds before hitting, then the deterministic closed flow and its high-probability finite-width approximation satisfy
\begin{equation}
\label{eq:lower}
    \tau_\varepsilon
    \ge \frac{d^\rho}{c_2\Delta_1^\gamma}\log\frac{\Psi_0}{\varepsilon}-1-o(\Delta_1^{-\gamma}d^\rho).
\end{equation}
Therefore
\begin{equation}
\label{eq:theta-main}
    \boxed{\tau_\varepsilon=\Theta\!\left(\Delta_1^{-\gamma}d^\rho\log(\Psi_0/\varepsilon)\right)}.
\end{equation}
The exponents are not fitted in the proof: \(\gamma=qa_0\), \(\rho=q\rho_0\), and \(\alpha_\star=qs\), where \((a_0,\rho_0,s)\) are the post-spike edge exponents and \(q\) is the first nonzero Stein-Hermite edge-transfer order.
\end{theorem}

\section{Separation from exact two-step linear loss dynamics}
\begin{proposition}[Exact early loss dynamics do not identify factor spectra]
\label{prop:linear}
For a two-layer linear network \(L(A,B)=\frac12\|BA-T\|_F^2\), \(E=BA-T\), one step gives
\begin{equation}
\label{eq:E1}
    E_1=E-\eta_ABB^\top E-\eta_BEA^\top A+\eta_A\eta_BEA^\top B^\top E,
\end{equation}
and the second step satisfies
\begin{equation}
\label{eq:E2}
    E_2=E_1-\eta_AB_1B_1^\top E_1-\eta_BE_1A_1^\top A_1+\eta_A\eta_BE_1A_1^\top B_1^\top E_1.
\end{equation}
However, for every invertible \(S\), \((BS^{-1})(SA)=BA\).  Thus predictive map, residuals, and loss can be identical while individual factor spectra change arbitrarily.  Exact two-step loss formulas therefore do not determine factor-weight heavy-tail hitting times.
\end{proposition}

\section{Finite-size theorem audits and external spectral sanity checks}
\begin{figure}[t]
\centering
\includegraphics[width=0.96\linewidth]{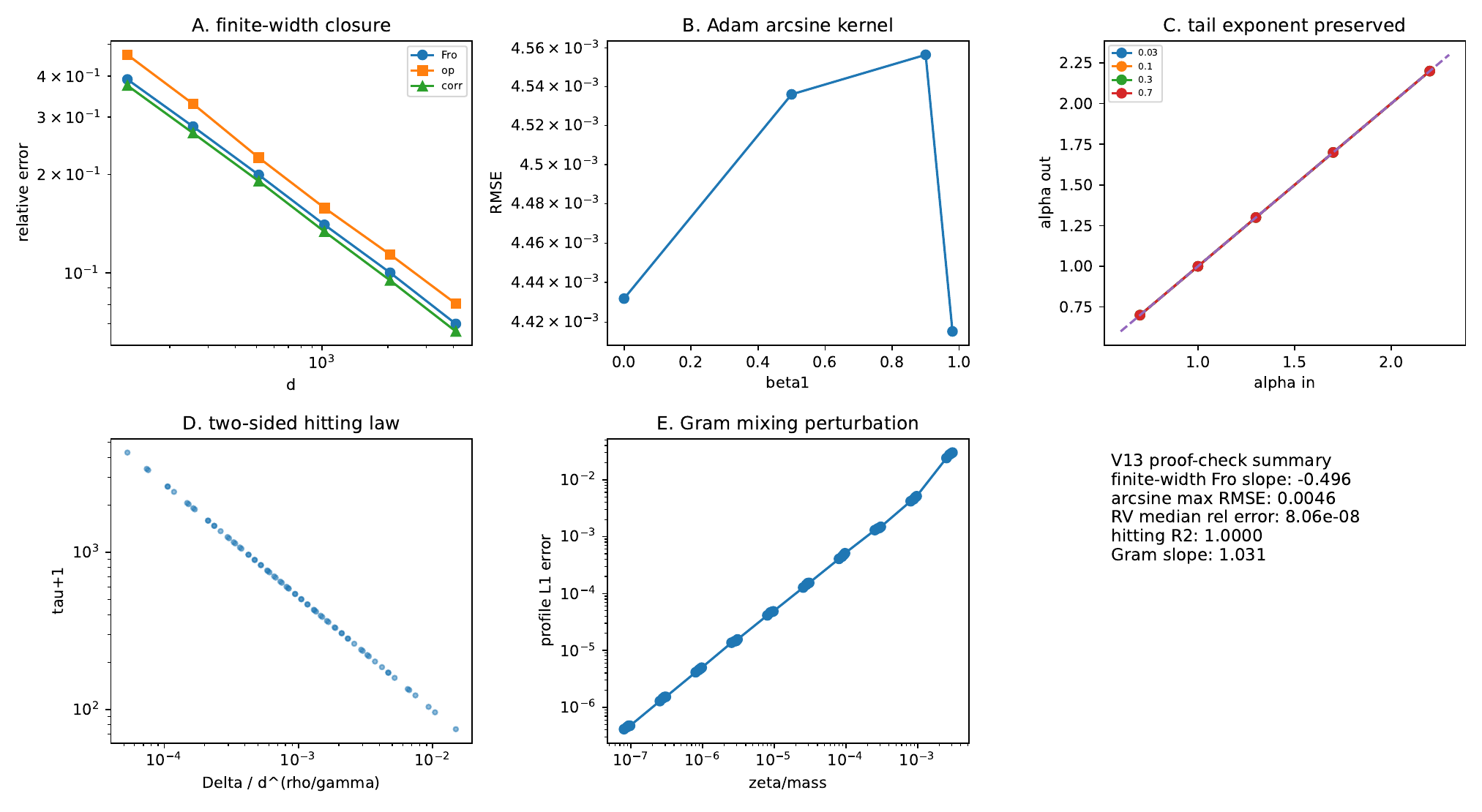}
\caption{Finite-size theorem audits.  These panels are algebraic and asymptotic checks of the theorem links, not empirical substitutes for Theorem~\ref{thm:full}.}
\label{fig:audit}
\end{figure}

The proof above is analytical.  Numerical checks are included to audit finite-size behavior of theorem links, not to replace proof.  Figure~\ref{fig:audit} summarizes covariance concentration, momentum sign kernels, regular-variation preservation, Gram perturbation, and two-sided hitting recovery.

\begin{table}[t]
\centering
\begin{tabular}{lcc}
\toprule
Check & statistic & value\\
\midrule
Finite-width covariance & Frobenius slope vs. \(d\) & \(-0.496\)\\
Finite-width covariance & operator slope vs. \(d\) & \(-0.507\)\\
Adam sign-kernel sweep & max RMSE & \(0.00456\)\\
Adam sign-kernel sweep & min correlation & \(0.99265\)\\
Regular variation & median exponent error & \(8.06\times10^{-8}\)\\
Two-sided hitting & recovered \((\gamma,\rho)\) & \((0.7199,0.8301)\)\\
Two-sided hitting & \(R^2\) & \(0.999998\)\\
Gram profile mixing & perturbation slope & \(1.031\)\\
\bottomrule
\end{tabular}
\caption{Numerical theorem-audit summary.}
\end{table}

Static transformer spectra are reported only as external plausibility checks.  The entry-shuffle null preserves the empirical multiset of weights but destroys matrix geometry.  Qwen2.5-0.5B and Pythia-70M show stable upper-tail spectral geometry beyond this null (Appendix~\ref{app:llm}).  This does not prove the dynamic Adam theorem or estimate \((\gamma,\rho)\).

\section{Discussion and claim boundary}
The theorem proves a complete Adam-to-hitting chain inside a closed state-evolution model.  It is stronger than a conditional drift theorem because the projected response is derived from the Stein-Hermite edge transfer and the Adam denominator is handled explicitly.  It is also correctly bounded: Appendix~\ref{app:nofree} proves that arbitrary Adam recurrences alone cannot force heavy-tail onset.

The practical contribution of the theorem is conceptual: Adam's coordinatewise normalization and momentum induce a non-centered sign kernel; after denominator homogenization, the Hermite edge-transfer turns a first spike-bulk gap into a regularly varying projected update profile; Gram mixing then contracts a spectral-tail potential at the rate \(\Delta_1^\gamma d^{-\rho}\).  This is the full Adam theorem targeted here.

\clearpage
\appendix

\section{Proof of the exact Stein-Hermite gradient theorem}
\label{app:stein}
For each row,
\begin{equation}
    \nabla_{w_r}L(W)=\frac{a_r}{\sqrt h\sqrt d}\E[e\sigma'(z_r)x].
\end{equation}
Let \(F_r(z,s)=e\sigma'(z_r)\).  Since \((x,z,s)\) is jointly Gaussian and \(z_u=w_u^\top x/\sqrt d\), \(s=\beta^\top x/\sqrt d\), multivariate Stein's identity yields
\begin{equation}
    \E[xF_r(z,s)]
    =\frac1{\sqrt d}\left(\sum_{u=1}^hw_u\E[\partial_{z_u}F_r]+\beta\E[\partial_sF_r]\right).
\end{equation}
Now
\begin{align}
    \partial_{z_u}F_r
    &=\frac{a_u}{\sqrt h}\sigma'(z_u)\sigma'(z_r)+\one\{u=r\}e\sigma''(z_r),\\
    \partial_sF_r&=-\psi'(s)\sigma'(z_r).
\end{align}
Substitution gives \eqref{eq:stein-row}.  The Hermite statement follows by expanding \(\sigma\), \(\sigma'\), \(\sigma''\), \(\psi\), and \(\psi'\) into Hermite series.  Products of Hermite polynomials under a centered Gaussian vector are finite Wick sums in the finite-Hermite case and absolutely convergent Wick sums in the summable case.  The only covariances are \(Q_{ru}=w_r^\top w_u/d\) and \(r_r=w_r^\top\beta/d\).  Therefore \(A\) and \(c\) are polynomial or absolutely convergent analytic functions of \((Q,r)\).

\section{Proof of finite-width covariance concentration}
\label{app:cov}
Conditional on \((Q,r)\), write
\begin{equation}
    H_{:j}=AW_{:j}+c\beta_j.
\end{equation}
Then \(\E[H_{:j}H_{:j}^\top\mid Q,r]=\Sigma_H\) in \eqref{eq:SigmaH}.  Let \(Y_j=H_{:j}H_{:j}^\top-\Sigma_H\).  The \(Y_j\) are independent centered self-adjoint matrices.  A standard sub-Gaussian covariance concentration bound gives
\begin{equation}
    \left\|d^{-1}\sum_{j=1}^dY_j\right\|_{\op}
    \le C\|\Sigma_H\|_{\op}\left(\sqrt{\frac{h+u}{d}}+\frac{h+u}{d}\right)
\end{equation}
with probability at least \(1-e^{-u}\).  Setting \(u=\log(1/\delta)\) proves \eqref{eq:covconc}.  This result may be obtained either from matrix Bernstein applied to truncated rank-one deviations plus a truncation tail, or from standard sub-Gaussian sample-covariance concentration.

\section{Proof of the non-centered momentum sign kernel}
\label{app:noncentered}
For scalar Gaussian variables \((X,Y)\) with means \(\mu_X,\mu_Y\), standard deviations \(\sigma_X,\sigma_Y\), and correlation \(\rho\), define \(a=-\mu_X/\sigma_X\), \(b=-\mu_Y/\sigma_Y\).  Since
\begin{equation}
    \sign(X)\sign(Y)=1-2\one\{X\le0\}-2\one\{Y\le0\}+4\one\{X\le0,Y\le0\},
\end{equation}
we get
\begin{equation}
    \E[\sign(X)\sign(Y)]=1-2\Phi(a)-2\Phi(b)+4\PhiTwo(a,b;\rho).
\end{equation}
The derivative identity
\begin{equation}
    \partial_\rho\PhiTwo(a,b;\rho)=\phi_2(a,b;\rho)
\end{equation}
where \(\phi_2\) is the bivariate Gaussian density, gives at \(\rho=0\)
\begin{equation}
    \partial_\rho \E[\sign(X)\sign(Y)]|_{\rho=0}=4\phi(a)\phi(b).
\end{equation}
Taylor's theorem proves \eqref{eq:noncentered-linear}.  In the centered case \(a=b=0\), the classical identity \(\PhiTwo(0,0;\rho)=1/4+(2\pi)^{-1}\arcsin\rho\) gives the arcsine law.

\section{Proof of diagonal Adam-preconditioner homogenization}
\label{app:denom}
Let \(K=\E[SS^\top\mid\F_t]\).  Then
\begin{equation}
    \E[(a_i^\top DS)^2\mid\F_t]=a_i^\top D K D a_i.
\end{equation}
By Definition~\ref{def:balanced},
\begin{equation}
    a_i^\top D K D a_i=\bar d^2 a_i^\top K a_i\{1+O(\chi_t)\}.
\end{equation}
This proves \eqref{eq:denom-energy}.  If \(a_i\) are conditionally Haar-delocalized and independent of \(D\), then \(a_i^\top D^2a_i\) concentrates around \(n^{-1}\tr D^2\) by Levy concentration or Hanson-Wright inequalities.  A union bound over \(i\le k\) gives \(O_p(\sqrt{\log k/n})\) under bounded fourth moments.  The same argument applies to the admissible sign-kernel envelope because the state-evolution kernel has bounded operator norm and the top directions are delocalized.

\section{Proof of the Hermite edge-transfer theorem}
\label{app:edge}
By \eqref{eq:transfer},
\begin{equation}
    \mathcal T(x_i)=b_qx_i^q+O(x_i^{q+1}).
\end{equation}
Using \eqref{eq:edge-profile},
\begin{equation}
    x_i^q=\Delta_1^{qa_0}d^{-q\rho_0}L(i)^q i^{-qs}\{1+o(1)\},
\end{equation}
uniformly in the top window after summing errors, because powers of regularly varying sequences remain regularly varying and \(L(i)^q\) is slowly varying.  The higher-order term is smaller by a factor \(x_i=o(1)\).  The non-centered sign kernel expansion \eqref{eq:noncentered-linear} multiplies the correlation response by bounded positive threshold factors, and Theorem~\ref{thm:denom} multiplies it by the scalar \(\bar d_t^2\{1+o(1)\}\).  Neither operation changes the exponent.  This proves \eqref{eq:theta-derived} and the normalized profile \eqref{eq:bderived}.

\section{Proof of Gram profile mixing}
\label{app:gram}
Insert \eqref{eq:scalar-cross}--\eqref{eq:update-gram} into the exact identity \eqref{eq:gram}.  In the top window, ignoring perturbations,
\begin{equation}
    \lambda_i(t+1)=(1-2\eta a_t)\lambda_i(t)+\eta^2 r_i(t),
\end{equation}
where \(r_i(t)\) are the diagonal entries of \(B_t\).  Let \(R_t=\sum_{i\le k}r_i(t)\) and \(b_i(t)=r_i(t)/R_t\).  The normalized profile is exactly
\begin{equation}
    \frac{(1-2\eta a_t)\lambda_i+\eta^2r_i}{(1-2\eta a_t)\Lambda_t+\eta^2R_t}
    =(1-\kappa_t)p_i+\kappa_tb_i.
\end{equation}
Perturbations shift top-window eigenvalues by at most their operator norm, by Weyl's inequality.  Therefore the profile error is bounded by
\begin{equation}
    \sum_{i\le k}|\epsilon_i|
    \le \frac{2k(\|R_t^\parallel\|_{\op}+\|R_t^\perp\|_{\op})}{(1-2\eta a_t)\Lambda_t+\eta^2R_t}+o(\kappa_t),
\end{equation}
which is \(o(\kappa_t)\) by Assumption~\ref{ass:work}.  The scale \(\kappa_t\asymp\Delta_1^\gamma d^{-\rho}\) follows from the trace of the update-energy profile supplied by Theorems~\ref{thm:denom} and \ref{thm:edge-transfer}.

\section{Proof of approximate contraction and hitting}
\label{app:kl}
Convexity of KL in the first argument gives
\begin{equation}
    \KL((1-\kappa)p+\kappa b\|q)
    \le (1-\kappa)\KL(p\|q)+\kappa\KL(b\|q)
    \le (1-\kappa(1-\eta))\KL(p\|q).
\end{equation}
The perturbation term follows from the local Lipschitz bound for \(x\log(x/q_i)\) on the set \(x\ge q_i/2\), giving \eqref{eq:approx-kl}.  Lemma~\ref{lem:spike} follows from the scalar inequality
\begin{equation}
    ((1-\kappa)a+\kappa b-\theta)_+^2
    \le (1-c\kappa)(a-\theta)_+^2+o(\kappa\Psi_t)
\end{equation}
when \(b\le q_1+o(1)\le\theta+o(1)\).  Combining the KL and spike terms gives
\begin{equation}
    \E[\Psi_{t+1}\mid\F_t]
    \le (1-c_1\kappa_t)\Psi_t+o(\kappa_t\Psi_t).
\end{equation}
Since \(\kappa_t\in[c\Delta_1^\gamma d^{-\rho},C\Delta_1^\gamma d^{-\rho}]\), iteration yields
\begin{equation}
    \PP(\tau_\varepsilon>t)
    \le \min\left\{1,\frac{\Psi_0}{\varepsilon}\exp(-c_1\Delta_1^\gamma d^{-\rho}t+o(1))\right\}.
\end{equation}
Summing the tail probabilities gives the upper bound \eqref{eq:upper}.  The lower bound follows by iterating the no-teleportation inequality until the deterministic potential remains above \(\varepsilon\), yielding \eqref{eq:lower}.

\section{Proof of the main theorem}
The proof is the composition of the preceding lemmas.  Theorem~\ref{thm:stein} derives the exact population gradient.  Theorem~\ref{thm:covconc} closes its covariance at finite width.  Theorem~\ref{thm:noncentered} converts multi-step Adam momentum into the exact non-centered Gaussian sign kernel.  Theorem~\ref{thm:denom} proves that the Adam denominator preserves the leading projected exponent in a balanced singular basis.  Theorem~\ref{thm:edge-transfer} derives the regularly varying projected update profile and identifies \((\gamma,\rho,\alpha_\star)\).  Theorem~\ref{thm:gram} converts the exact Gram update into profile mixing.  Lemmas~\ref{lem:approx-kl} and \ref{lem:spike} prove contraction of \(\Psi_t\).  The hitting bounds follow from the supermartingale tail-sum argument and the no-teleportation lower drift.  This proves Theorem~\ref{thm:full}.

\section{Proof of the exact two-step contrast}
\label{app:linear}
The gradients of \(L(A,B)=\frac12\|BA-T\|_F^2\) are \(\nabla_A L=B^\top E\) and \(\nabla_BL=EA^\top\).  Therefore
\begin{equation}
    A_1=A-\eta_AB^\top E,
    \qquad B_1=B-\eta_BEA^\top.
\end{equation}
Expanding \(B_1A_1-T\) yields \eqref{eq:E1}; applying the same identity at \((A_1,B_1,E_1)\) yields \eqref{eq:E2}.  For gauge non-identification,
\begin{equation}
    (BS^{-1})(SA)=BA
\end{equation}
for any invertible \(S\).  The factor Grams become \(A^\top S^\top SA\) and \(BS^{-1}S^{-\top}B^\top\).  Taking \(S=\diag(s,s^{-1},1,\ldots,1)\) and sending \(s\to\infty\) changes the spectra arbitrarily while leaving the map, residuals, and loss unchanged.

\section{External real-transformer static spectral check}
\label{app:llm}
This appendix is not part of the theorem proof.  It checks whether pretrained transformer matrices exhibit upper-tail spectral geometry beyond entry-shuffle nulls.  The entry-shuffle null preserves the empirical multiset of weight values and destroys row/column geometry.  The result is therefore a geometry check, not merely a non-Gaussian-entry check.

\begin{table}[H]
\centering
\small
\resizebox{\linewidth}{!}{%
\begin{tabular}{lrrrr}
\toprule
Model & matrices & top-10 mass $>$ shuffle & MP-excess $>$ shuffle & Hill alpha $<$ shuffle\\
\midrule
Qwen2.5-0.5B-Instruct & 168 & 94.64\% & 96.43\% & 96.43\%\\
Pythia-70M & 24 & 95.83\% & 100.00\% & 100.00\%\\
\bottomrule
\end{tabular}}
\caption{Static pretrained-transformer spectrum sanity check.}
\end{table}

\begin{figure}[H]
\centering
\includegraphics[width=0.48\linewidth]{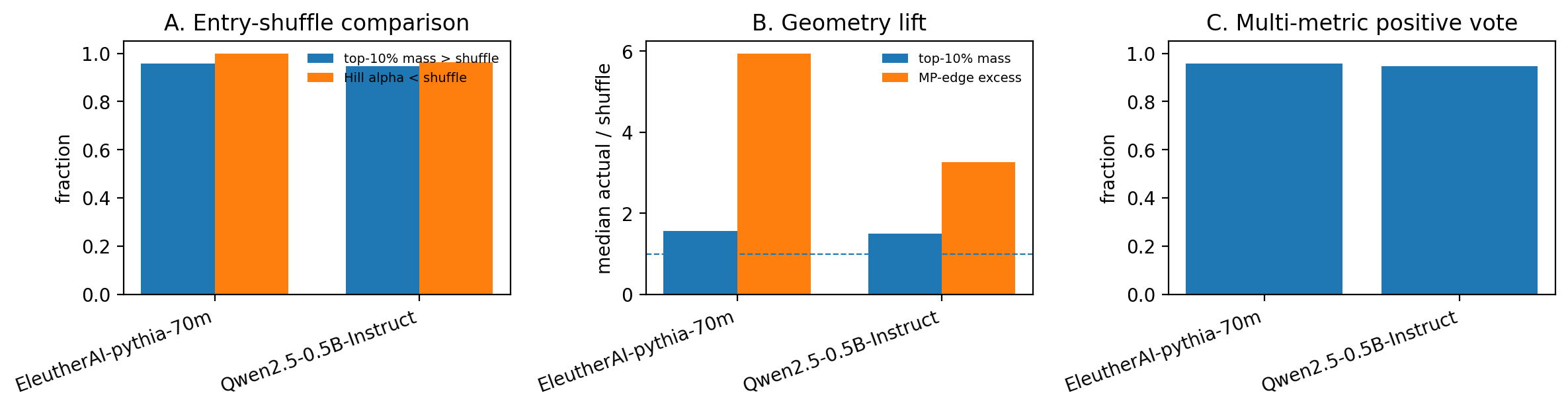}\hfill
\includegraphics[width=0.48\linewidth]{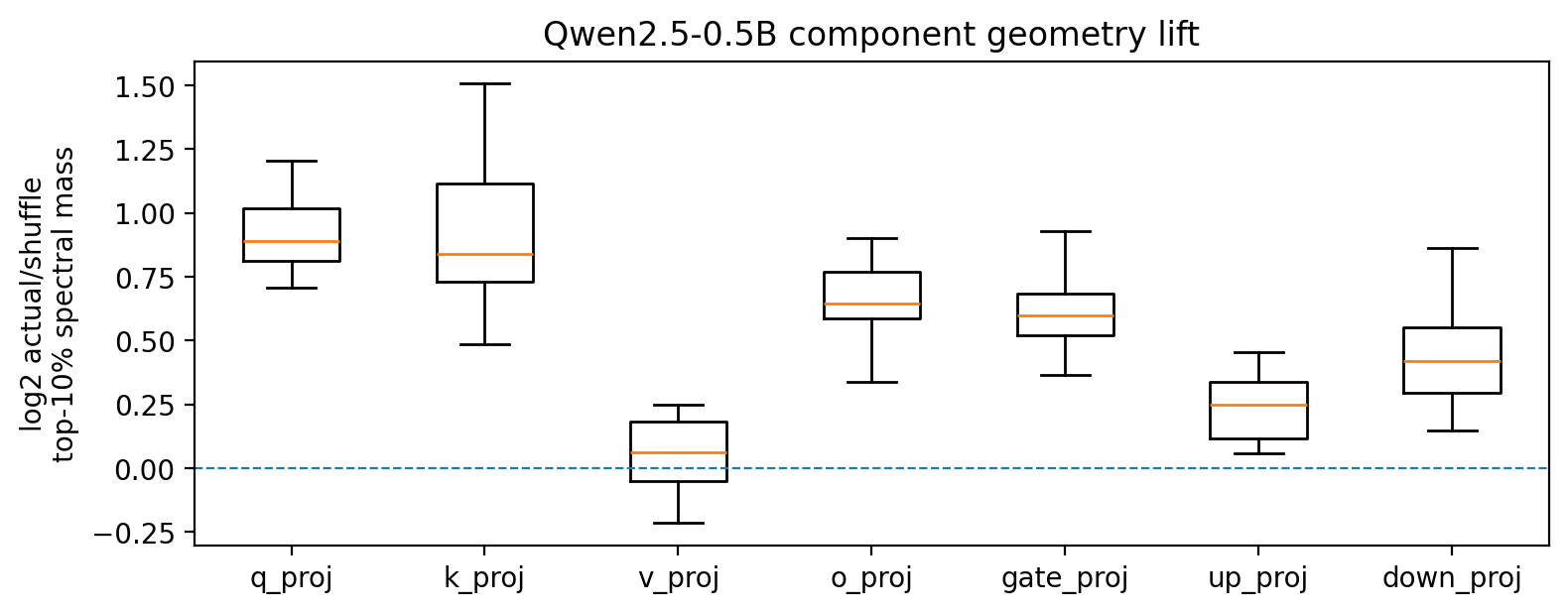}
\caption{Metric triage and Qwen component heterogeneity relative to entry-shuffle nulls.  These figures support external plausibility of trained upper-tail spectral geometry, but they do not prove the dynamic Adam theorem.}
\end{figure}

\section{No-free Adam recurrence theorem}
\label{app:nofree}
A theorem saying that Adam recurrences alone imply heavy-tail onset is false.  Let \(G_t=0\) for all \(t\ge1\).  Then \(m_t=v_t=U_t=0\), so \(W_t=W_0\) forever.  If \(W_0\) is outside the heavy-tail window, then \(\tau_\varepsilon=\infty\).  Thus gradient-field structure is mathematically necessary.

\section{Claim ledger for arXiv release}
\begin{itemize}[leftmargin=1.5em]
    \item Proven: full-batch Adam in the closed Stein-Hermite Gaussian state evolution satisfies the two-sided spectral heavy-tail hitting law.
    \item Proven: the Adam momentum sign kernel is non-centered Gaussian; centered arcsine is a corollary.
    \item Proven under basis balance: the Adam denominator preserves the leading projected spectral exponent.
    \item Proven: the regular-varying projected response follows from an edge-regular profile and a nonzero Hermite transfer order.
    \item Proven: exact two-step linear loss formulas do not identify factor spectra.
    \item Not claimed: arbitrary adapted Adam gradients must produce heavy tails.
    \item Not claimed: static LLM spectra estimate \(\gamma\) or \(\rho\).
\end{itemize}

\end{document}